\documentclass[conference]{IEEEtran}
\IEEEoverridecommandlockouts
\usepackage{cite}
\usepackage{amsmath,amssymb,amsfonts}
\usepackage{graphicx}
\usepackage{textcomp}
\usepackage{xcolor}
\usepackage{calligra}
\usepackage{algpseudocode} 
\usepackage{amsmath} 
\usepackage{booktabs}  
\usepackage{multirow}  
\usepackage{longtable}  

\def\BibTeX{{\rm B\kern-.05em{\sc i\kern-.025em b}\kern-.08em
    T\kern-.1667em\lower.7ex\hbox{E}\kern-.125emX}}
\begin{document}

\title{FedGIG: Graph Inversion from Gradient 
in Federated Learning}

\author{
\IEEEauthorblockN{
Tianzhe Xiao,
Yichen Li,
Yining Qi,
Haozhao Wang, and
Ruixuan Li}
\IEEEauthorblockA{School of Computer Science and Technology, Huazhong University of Science and Technology, Wuhan, China}
\IEEEauthorblockA{\{tz\_\_xiao, ycli0204, qiyining, hz\_wang, rxli\}@hust.edu.cn}
}

\maketitle

\begin{abstract}
Recent studies have shown that Federated learning (FL) is vulnerable to Gradient Inversion Attacks (GIA), which can recover private training data from shared gradients. However, existing methods are designed for dense, continuous data such as images or vectorized texts, and cannot be directly applied to sparse and discrete graph data. This paper first explores GIA's impact on Federated Graph Learning (FGL) and introduces Graph Inversion from Gradient in
Federated Learning (FedGIG), a novel GIA method specifically designed for graph-structured data. FedGIG includes the adjacency matrix constraining module, which ensures the sparsity and discreteness of the reconstructed graph data, and the subgraph reconstruction module, which is designed to complete missing common subgraph structures. Extensive experiments on molecular datasets demonstrate FedGIG's superior accuracy over existing GIA techniques.  
\end{abstract}


\section{Introduction}

Federated Learning (\textbf{FL}) has become a popular privacy-preserving approach to distributed machine learning\cite{Mcmahan17,li2024unleashing,wang2024fedcda,li2024rehearsal,li_tpds,Li_2024_CVPR}. It enables multiple parties to collaboratively train a shared model by exchanging only model updates (gradients), rather than raw data. Due to the benefits of preserving privacy and communication efficiency, FL has been widely deployed in various applications, such as smart healthcare \cite{Antunes22,Rahman2023} and finance analysis \cite{Yang19,Long2020}. While FL prevents the direct sharing of raw data, recent studies shows that the gradients exchanged between clients during training can still reveal crucial information about the underlying data, meaning that FL remains vulnerable to Gradient Inversion Attacks \textbf{(GIA)}.

DLG\cite{Zhu19} is currently one of the most widely used GIA baselines. It simultaneously optimizes a randomly generated pseudo training dataset and labels, with the objective of minimizing the difference between the gradients computed from the model's output using these pseudo data and the real gradients. This approach enables clear reconstruction of images from training set like CIFAR-100, even some vectorized texts. IDLG\cite{Zhao20}, building upon DLG, enhances the attack's accuracy by using the sign of the gradients from the linear layer to infer the true labels for image classification tasks. Subsequent works have introduced regularization terms that encourage smoother reconstructed images\cite{Geiping20}, allowing for clearer image recovery under larger models or bigger batch sizes\cite{Yin21,huang2021evaluating,zhang2022survey}.


Although the aforementioned methods are useful, they are designed for FL systems that use image and text data. However, with the development of graph data, Federated Graph Learning (\textbf{FGL}) has become the mainstream approach. As an extension of FL that specifically targets graph-structured data\cite{chen2024fedgl}, it’s being widely used in privacy-sensitive domains such as healthcare, social networks, and financial services. Existing attack methods are not suitable for FGL for the following reasons: (i) graph data is discrete. Existing methods designed for traditional Euclidean data types like images or vectorized natural language, require the optimized data to be continuously derivable to provide the direction of data update.  In contrast, for graph data such as molecular graphs or social networks, the adjacency matrix representing the graph structure is typically composed of discrete Boolean variables, which makes the data non-differentiable, causing traditional GIA methods to fail to converge during the optimization process.
(ii) graph data is typically sparse and may possess unique intrinsic properties, such as symmetry. This also contrasts sharply with the dense image data and vectorized text data that existing GIA methods are designed to handle. Therefore, it is urgent to improve the existing GIA method around the characteristics of the graph data so as to accurately restore the sparse and discrete graph data that cannot be directly differentiated, and study the real vulnerability of GFL when facing GIA.

In this paper, we propose the first GIA method specifically designed for FGL, which we call \textbf{FedGIG}. Our approach is explicitly designed to target the unique gradient inversion vulnerabilities in FGL by incorporating the inherent structural characteristics of graph data. Specifically, our method leverages the \textit{discreteness}, \textit{sparsity}, and \textit{symmetry} of graph structures to guide the optimization process, making the attack more efficient and accurate. We introduce an optimization framework that keeps the adjacency matrix under constraint and iteratively prunes unnecessary edges during the reconstruction of the graph, which helps refine the accuracy of the graph inversion. Additionally, to address the challenge of reconstructing complex local graph patterns, we propose using a masked graph autoencoder, which aids in recovering the specific local structures that commonly occur in graph data.

By focusing on the unique challenges posed by graph data, we reveal the extent to which FGL is susceptible to gradient inversion attacks and provide a deeper understanding of the privacy risks associated with FGL in practice. Our main contributions are summarized as follows:
\begin{itemize}
    \item We introduce the novel FedGIG attack method, specifically designed to target the gradient inversion vulnerabilities inherent in FGL.
    \item We exploit the inherent structural properties of graph data, such as discreteness, sparsity, and symmetry, to optimize the reconstruction of graph structures and propose the adjacency matrix constraining module, which ensures the sparsity and discreteness of the reconstructed graph data, and the subgraph reconstruction module, which can complete missing common subgraph structures.
    \item We demonstrate the effectiveness of our attack method through extensive experiments on five benchmark molecular graph datasets, showing that our approach outperforms existing gradient inversion attack techniques in terms of accuracy and efficiency when applied to graph-based models in FGL.
\end{itemize}

\section{Preliminaries}

In this section, we briefly introduce the key concepts of federated graph learning and gradient inversion attack in FGL, which are fundamental in understanding the design rationale and workflow of our proposed FedGIG method.

\subsection{Federated Graph Learning}

FL enables multiple clients to collaboratively train a shared model while maintaining data privacy by only exchanging model updates (gradients), rather than raw data. FGL extends FL to graph data, where each client computes local gradients $\nabla_{\mathbf{W}^{k}} \mathcal{L}(\mathbf{W}^{k})$ based on its local graph data $\mathcal{G}^{k} = (\mathcal{V}^{k}, \mathcal{A}^{k}, \mathbf{X}^{k})$. Here, $\mathcal{V}^{k}$ represents the set of nodes in the $k$-th client’s graph, $\mathcal{A}^{k}$ denotes the adjacency matrix, and $\mathbf{X}^{k}$ is the node feature matrix for the graph. The global dataset is considered as the composition of all local graph datasets, $\mathcal{G} = \sum_{k=1}^{K} \mathcal{G}^{k}$. The objective of FGL is to learn a global model $w$ that minimizes the total empirical loss over the entire dataset $\mathcal{G}$:
\begin{equation}
\min_{w} \mathcal{L}(w) := \sum_{k=1}^{K} \frac{|\mathcal{G}^{k}|}{|\mathcal{G}|} \mathcal{L}^{k}(w).
\end{equation}
where $\mathcal{L}^{k}(w)$ is the local loss in the $k$-th client, which is calculated based on the local graph data $\mathcal{G}^{k}$.

\subsection{Gradient Inversion Attack in FGL}

In Federated Graph Learning, GIA aims to reconstruct sensitive graph data by exploiting the gradients $\nabla_{w^{k}}$ exchanged between clients and the server. The attack focuses on minimizing the loss between the true gradients $\nabla_{w^{k}}$ and the gradients $\hat{\nabla_{w^{k}}}$ from a reconstructed graph, $\hat{\mathcal{G}}^{k} = (\hat{\mathcal{V}}^{k}, \hat{\mathcal{A}}^{k}, \hat{\mathbf{X}}^{k})$. The objective of the gradient inversion attack is to solve the following optimization problem:
\begin{equation}
\hat{\mathcal{G}}^{k} = \arg \min_{\hat{\mathcal{G}}^{k}} \left\| \nabla_{w^{k}} \mathcal{L}(\hat{\mathcal{G}}^{k}) - \nabla_{w^{k}} \mathcal{L}(\mathcal{G}^{k}) \right\|^2.
\end{equation}
This reconstruction process aims to recover node features $\hat{\mathbf{X}}^{k}$ and edge relationships $\hat{\mathcal{A}}^{k}$ by leveraging the gradients exchanged during training. Our focus in this work is on optimizing the reconstruction of $\mathcal{A}^{k}$ to enhance the accuracy of GIA.

\begin{figure*}[t]
\begin{center}
\centerline{\includegraphics[width=\linewidth]{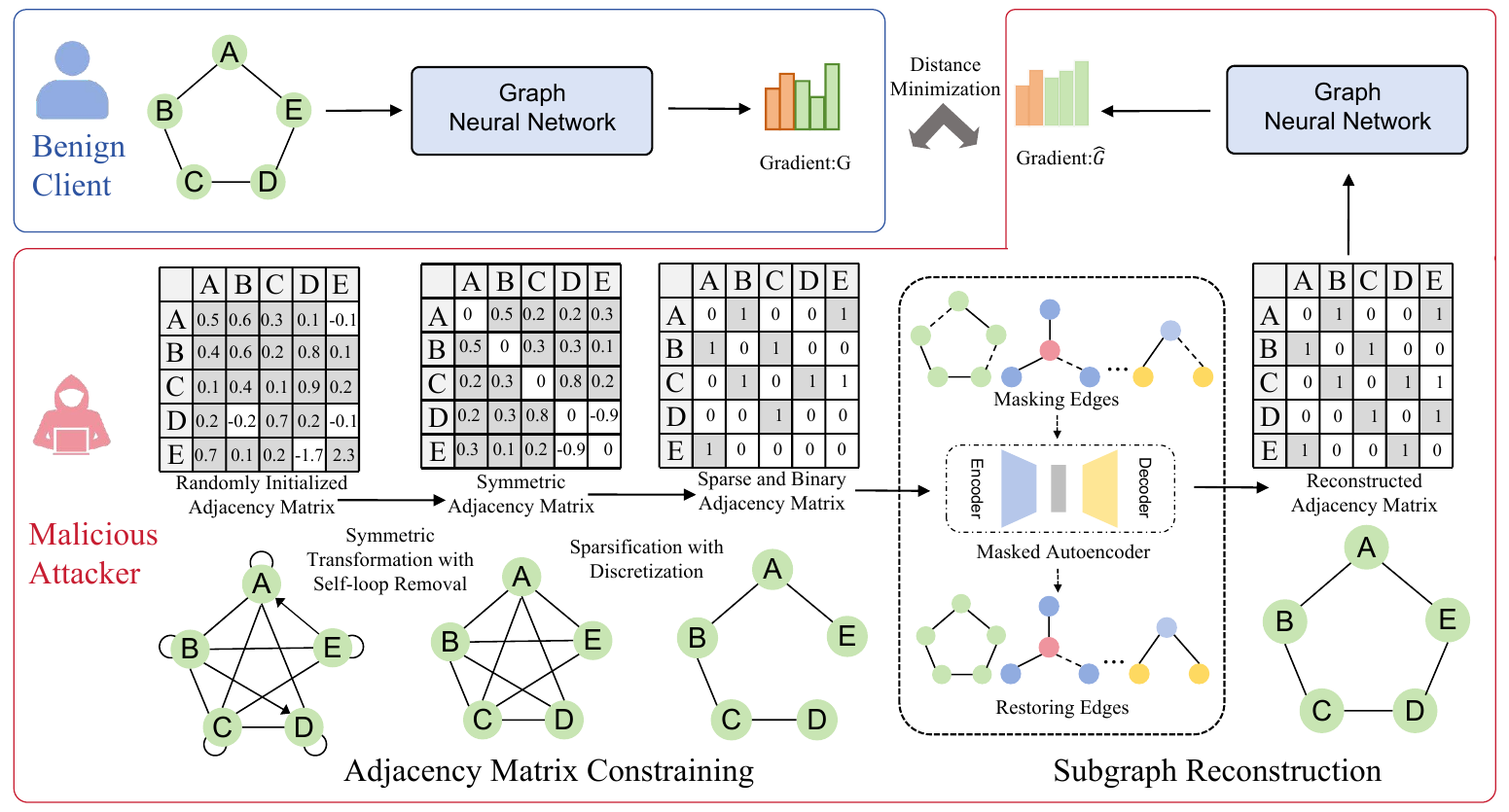}}
\caption{The overall architecture of FedGIG. A benign client in FGL will use its local graph data to compute gradients updating to the server, while honest but curious malicious attackers can initialize a pseudo graph data, using the adjacent matrix constraining and subgraph reconstruction modules in FedGIG to accelerate the attack and help the pseudo graph data to be optimized fairly close to the real client’s data when the distance between the calculated gradients is small enough if the real data and the pseudo data are fed into the same GNN network.
}
\label{framework}
\end{center}
\end{figure*}

\section{Related Work}
\noindent\textbf{Federated Graph Learning (FGL).} FGL addresses the challenges of training graph neural networks across distributed data silos while preserving data privacy. Research has primarily focused on several key directions: techniques to handle data heterogeneity across clients \cite{chen2024fedgl}, including personalized model training and data normalization methods, privacy-preserving mechanisms \cite{wu2022multi,wu2021fedgnn,wu2022federated}, including the use of differential privacy, secure multi-party computation, and homomorphic encryption, strategies to reduce communication costs \cite{chen2021fedgraph,yao2024fedgcn}, where researchers have explored strategies to reduce the overhead, such as model compression, sparsification, and efficient aggregation algorithms, methods to enhance model robustness and fairness \cite{pan2024towards,Wang_2023_CVPR},  and the development of frameworks supporting diverse GNN models \cite{wang2022federatedscope,he2021fedgraphnn}. However, no existing works focus on gradient inversion attacks against FGL.

\noindent\textbf{Gradient Inversion Attack(GIA).}
Federated learning, while offering a promising approach for distributed training, remains vulnerable to various malicious attacks, including gradient inversion \cite{wei2020framework}, model inversion \cite{li2022ressfl}, and membership inference \cite{zhang2020gan} attacks. Among these, gradient inversion poses a particularly severe threat as it can expose sensitive information from the victim's training data. This attack technique involves initializing artificial training data and labels, which are then optimized to mimic the gradients of the real data. As the artificial and actual gradients align, the synthetic data starts to mirror the characteristics of the victim's private data. The attack methodology outlined in \cite{Zhu19} has been proven effective against both computer vision and natural language processing tasks. Subsequent advancements have focused on enhancing the attack's efficacy through the use of prior knowledge for initialization \cite{Jeon21}, extraction of ground-truth labels \cite{Zhao20}, accelerated optimization \cite{Geiping20}, and the incorporation of regularization terms \cite{Geiping20,Yin21}. These refinements have enabled GIA to be more potent, targeting larger batch sizes, higher resolution data, and more sophisticated models, such as Vision Transformers (ViT) \cite{Yin21}. There is no existing work to optimize GIA specifically for the characteristics of the graph.

\section{Proposed Method}

In this section, we introduce our proposed method, FedGIG, to enhance the accuracy of graph structure recovery in GIA. The method consists of two main parts: the adjacency matrix constraining module that ensures the sparsity and discreteness of the reconstructed graph data, and the subgraph reconstruction module that can the complete missing common subgraph structures, the overall architecture of FedGIG is illustrated in Figure \ref{framework}.

\subsection{Adjacency Matrix Constraining}

The adjacency matrix constraining process includes two steps: symmetric transformation with self-loop removal and sparsification with discretization. Let $\mathbf{A}_i$ represent the adjacency matrix of the graph from the $i$-th client in a Federated Learning (FL) setting.

\subsubsection{Symmetric Transformation with Self-loop Removal}

Since most graph data in FGL, such as chemical molecular structures and social networks, are represented by undirected graphs, we focus on ensuring symmetry and removing self-loops during the reconstruction process. The adjacency matrix $\mathbf{A}^{k}$ can be transformed in two steps: (i) remove the self-loops and (ii) symmetrize the adjacency matrix to ensure undirectedness.

We perform a transformation to $\mathbf{A}^{k}$ to simultaneously apply both self-loop removal and symmetric transformation. The transformation is defined as:
\begin{equation}
\mathbf{A}^{k}_{\text{sym}} = \frac{1}{2} \left( (\mathbf{A}^{k} - \text{diag}(\mathbf{A}^{k})) + (\mathbf{A}^{k} - \text{diag}(\mathbf{A}^{k}))^\top \right).
\end{equation}
where $\text{diag}(\mathbf{A}^{k})$ denotes the diagonal matrix formed by extracting the diagonal entries of $\mathbf{A}^{k}$, effectively removing the self-loops by setting the diagonal elements to zero. Then, the matrix is symmetrized by averaging it with its transpose, ensuring that the adjacency matrix remains undirected. This transformation ensures that $\mathbf{A}^{k}_{\text{sym}}$ represents an undirected graph with no self-loops, so the adjacency matrix in the attack is promptly normalized to prevent deviations during the optimization process.

\subsubsection{Sparsification with Discretization}

After symmetrizing the adjacency matrix, we perform the sparsification step, where we select the top $n(t)$ edges for each node based on the sigmoid-transformed values of the adjacency matrix $\mathbf{A}^{k}_{\text{sym}}$. The number of selected edges, $n(t)$, decays over time to ensure smooth convergence. This process can be mathematically described as follows.

We first apply the sigmoid function to the symmetrized adjacency matrix $\mathbf{A}^{k}_{\text{sym}}$, and then select the top $n(t)$ largest values for each node, setting all other values to zero. This operation can be mathematically represented as:

\begin{equation}
\mathbf{A}^{k}_{\text{tmp}}(i,j) = 
\begin{cases}
\sigma(\mathbf{A}^{k}_{\text{sym}}(i,j)), & \text{if} \ j \in \mathcal{I}_{n(t)}(\sigma(\mathbf{A}^{k}_{\text{sym},i,:})), \\
0, & \text{otherwise}.
\end{cases}
\end{equation}
Here, $\sigma(\mathbf{A}^{k}_{\text{sym}}(i,j))$ applies the sigmoid function element-wise to the symmetrized adjacency matrix $\mathbf{A}^{k}_{\text{sym}}$, producing a probability matrix. $\mathcal{I}_{n(t)}(\sigma(\mathbf{A}^{k}_{\text{sym},i,:}))$ is the index set of the top $n(t)$ largest values in the $i$-th row of $\sigma(\mathbf{A}^{k}_{\text{sym}})$. The elements corresponding to these top $n(t)$ largest values are retained, and all other values are set to zero. The number of edges selected, $n(t)$, decays over time as follows:

\begin{equation}
n(t) = \max(n_0 - \alpha t, n_{\max}).
\end{equation}
where $n_0$ is the initial number of edges, $\alpha$ is the decay rate, and $n_{\max}$ is the maximum number of edges finally allowed. This decay function ensures that as the optimization progresses, fewer edges are selected, promoting sparsity.

Next, we perform binary discretization by comparing the transformed values to a threshold $\beta$. The final binary adjacency matrix $\mathbf{A}^{k}_{\text{binary}}$ is then obtained by applying the threshold function:

\begin{equation}
\mathbf{A}^{k}_{\text{binary}}(i,j) =
\begin{cases}
1, & \text{if} \ \mathbf{A}^{k}_{\text{tmp}}(i,j) \geq \beta, \\
0, & \text{otherwise}.
\end{cases}
\end{equation}
Here, $\beta$ is the binarization threshold. The edges whose transformed values are greater than or equal to $\beta$ are retained, and all other edges are discarded.

Finally, to enforce sparsity in the adjacency matrix, we introduce a regularization term that penalizes the total number of edges:

\begin{equation}
\mathcal{L}_{\text{total}} = \mathcal{L}_{\text{recon}} + \lambda \sum_{i,j} \mathbf{A}^{k}_{\text{binary}}(i,j).
\end{equation}
where $\lambda$ controls the strength of the sparsity penalty.

\subsection{Subgraph Reconstruction}

In graph structures, such as chemical molecules or protein graphs, local patterns like cycles and chains are critical for accurate reconstruction. However, gradient inversion attacks may fail to capture these patterns effectively. To address this issue, we employ Masked Graph Autoencoders (MGAE), which help reconstruct these local patterns. The process begins by randomly masking edges in the adjacency matrix $\mathbf{A}^k$, creating the masked adjacency matrix $\mathbf{A}^{k}_{\text{masked}}$. The goal is to learn a latent representation of the graph that enables accurate reconstruction of the original adjacency matrix. The encoder is a 2-layer Graph Convolutional Network (GCN), which takes the masked adjacency matrix $\mathbf{A}^{k}_{\text{masked}}$ and the node feature matrix $\mathbf{X}^k$ to produce node embeddings $\mathbf{Z}^k$:

\begin{equation}
\mathbf{Z}^k = \text{Encoder}(\mathbf{A}^{k}_{\text{masked}}, \mathbf{X}^k).
\end{equation}
These embeddings capture the graph's structural information. The decoder then reconstructs the adjacency matrix using the node embeddings $\mathbf{Z}^k$:

\begin{equation}
\hat{\mathbf{A}}^k = \sigma(\mathbf{Z}^k (\mathbf{Z}^k)^T).
\end{equation}
To compute the reconstruction loss, we use the Mean Squared Error between the original and reconstructed adjacency matrices. The  loss for each element $a_{ij}$ in the adjacency matrix is calculated as follows:
\begin{equation}
\mathcal{L}_{reconstruction} = \frac{1}{N^2} \sum_{i=1}^{N} \sum_{j=1}^{N} (a_{ij} - \hat{a}_{ij})^2.
\end{equation}
where $a_{ij}$ is the element in the original adjacency matrix $\mathbf{A}^k$, and $\hat{a}_{ij}$ is the corresponding element in the reconstructed adjacency matrix $\hat{\mathbf{A}}^k$. The term $(a_{ij} - \hat{a}_{ij})^2$ computes the squared difference between the original and reconstructed elements and the summation over all elements in the matrix, followed by the division by $N^2$, gives the average squared error.

Finally, the optimized adjacency matrix $\hat{\mathbf{A}}^k$ is obtained by passing the optimized binary adjacency matrix $\mathbf{A}^{k}_{\text{binary}}$ and node features $\mathbf{X}^k$ through the trained encoder-decoder network.
This method ensures that the reconstructed adjacency matrix captures both the global and local graph structures, improving the robustness of the reconstruction process.

\begin{table}[ht]
\caption{Comparison of DLG, DLG+BP, iDLG, iDLG+BP, and FedGIG (ours) on different datasets.}
\centering
\renewcommand{\arraystretch}{1.0} 
\begin{tabular}{l@{\hskip 5pt}c@{\hskip 5pt}p{1.4cm}@{\hskip 5pt}p{1.4cm}@{\hskip 5pt}p{1.4cm}@{\hskip 5pt}p{1.4cm}} 
\toprule
\multirow{2}{*}{\textbf{Dataset}} & \multirow{2}{*}{\textbf{Method}} & \multicolumn{4}{c}{\textbf{Metrics}} \\
\cmidrule(r){3-6} 
& & Accuracy $\uparrow$ & Jaccard $\uparrow$ & MSE $\downarrow$ & AUC $\uparrow$ \\
\midrule
\multirow{5}{*}{\textbf{MUTAG}} & DLG & 0.000 \tiny{$\pm$0.000} & 0.004 \tiny{$\pm$0.003} & 1.216 \tiny{$\pm$0.011} & 0.459 \tiny{$\pm$0.002} \\
& DLG+BP & 0.654 \tiny{$\pm$0.002} & 0.099 \tiny{$\pm$0.002} & 0.346 \tiny{$\pm$0.011} & 0.499 \tiny{$\pm$0.001} \\
& iDLG & 0.000 \tiny{$\pm$0.000} & 0.007 \tiny{$\pm$0.003} & 1.200 \tiny{$\pm$0.001} & 0.460 \tiny{$\pm$0.012} \\
& iDLG+BP & 0.638 \tiny{$\pm$0.006} & 0.118 \tiny{$\pm$0.005} & 0.314 \tiny{$\pm$0.015} & 0.512 \tiny{$\pm$0.001} \\
& FedGIG & \textbf{0.906} \tiny{$\pm$0.011} & \textbf{0.452} \tiny{$\pm$0.002} & \textbf{0.154} \tiny{$\pm$0.020} & \textbf{0.676} \tiny{$\pm$0.021} \\
\midrule
\multirow{5}{*}{\textbf{NCI1}} & DLG & 0.000 \tiny{$\pm$0.000} & 0.008 \tiny{$\pm$0.001} & 0.975 \tiny{$\pm$0.015} & 0.419 \tiny{$\pm$0.002} \\
& DLG+BP & 0.550 \tiny{$\pm$0.002} & 0.104 \tiny{$\pm$0.002} & 0.349 \tiny{$\pm$0.011} & 0.451 \tiny{$\pm$0.001} \\
& iDLG & 0.000 \tiny{$\pm$0.000} & 0.010 \tiny{$\pm$0.001} & 0.933 \tiny{$\pm$0.005} & 0.425 \tiny{$\pm$0.001} \\
& iDLG+BP & 0.555 \tiny{$\pm$0.002} & 0.100 \tiny{$\pm$0.002} & 0.323 \tiny{$\pm$0.015} & 0.453 \tiny{$\pm$0.001}  \\
& FedGIG & \textbf{0.936} \tiny{$\pm$0.002} & \textbf{0.463} \tiny{$\pm$0.012} & \textbf{0.103} \tiny{$\pm$0.002} & \textbf{0.679} \tiny{$\pm$0.013} \\
\midrule
\multirow{5}{*}{\textbf{Tox21}} 
& DLG & 0.000 \tiny{$\pm$0.000} & 0.068 \tiny{$\pm$0.003} & 1.014 \tiny{$\pm$0.001} & 0.399 \tiny{$\pm$0.002} \\
& DLG+BP & 0.583 \tiny{$\pm$0.007} & 0.083 \tiny{$\pm$0.001} & 0.397 \tiny{$\pm$0.021} & 0.459 \tiny{$\pm$0.002} \\
& iDLG & 0.000 \tiny{$\pm$0.000} & 0.075 \tiny{$\pm$0.001} & 0.948 \tiny{$\pm$0.002} & 0.433 \tiny{$\pm$0.010} \\
& iDLG+BP & 0.603 \tiny{$\pm$0.005} & 0.122 \tiny{$\pm$0.007} & 0.245 \tiny{$\pm$0.005} & 0.463 \tiny{$\pm$0.001} \\
& FedGIG & \textbf{0.893} \tiny{$\pm$0.002} & \textbf{0.434} \tiny{$\pm$0.001} & \textbf{0.132} \tiny{$\pm$0.007} & \textbf{0.655} \tiny{$\pm$0.011} \\
\midrule
\multirow{5}{*}{\textbf{Clintox}} 
& DLG & 0.000 \tiny{$\pm$0.000} & 0.006 \tiny{$\pm$0.001} & 1.001 \tiny{$\pm$0.002} & 0.414 \tiny{$\pm$0.001} \\
& DLG+BP & 0.593 \tiny{$\pm$0.003} & 0.045 \tiny{$\pm$0.001} & 0.352 \tiny{$\pm$0.004} & 0.463 \tiny{$\pm$0.002} \\
& iDLG & 0.000 \tiny{$\pm$0.000} & 0.007 \tiny{$\pm$0.001} & 1.000 \tiny{$\pm$0.001} & 0.418 \tiny{$\pm$0.002} \\
& iDLG+BP & 0.590 \tiny{$\pm$0.002} & 0.046 \tiny{$\pm$0.001} & 0.350 \tiny{$\pm$0.002} & 0.467 \tiny{$\pm$0.001} \\
& FedGIG & \textbf{0.914} \tiny{$\pm$0.003} & \textbf{0.487} \tiny{$\pm$0.001} & \textbf{0.114} \tiny{$\pm$0.002} & \textbf{0.682} \tiny{$\pm$0.005} \\
\midrule
\multirow{5}{*}{\textbf{FreeSolv}} 
& DLG & 0.000 \tiny{$\pm$0.000} & 0.025 \tiny{$\pm$0.002} & 1.032 \tiny{$\pm$0.003} & 0.454 \tiny{$\pm$0.001} \\
& DLG+BP & 0.645 \tiny{$\pm$0.002} & 0.053 \tiny{$\pm$0.001} & 0.356 \tiny{$\pm$0.005} & 0.469 \tiny{$\pm$0.001} \\
& iDLG & 0.000 \tiny{$\pm$0.000} & 0.024 \tiny{$\pm$0.002} & 1.021 \tiny{$\pm$0.002} & 0.460 \tiny{$\pm$0.002} \\
& iDLG+BP & 0.647 \tiny{$\pm$0.005} & 0.054 \tiny{$\pm$0.002} & 0.355 \tiny{$\pm$0.005} & 0.465 \tiny{$\pm$0.002} \\
& FedGIG & \textbf{0.943} \tiny{$\pm$0.002} & \textbf{0.556} \tiny{$\pm$0.001} & \textbf{0.088} \tiny{$\pm$0.001} & \textbf{0.755} \tiny{$\pm$0.003} \\
\bottomrule
\end{tabular}
\label{table1}
\end{table}

\section{Experiments}
\subsection{Experimental Setups}
\noindent\textbf{Metrics.} We use four metrics to measure the similarity between the reconstructed adjacency matrix and the original adjacency matrix after the GIA attack: Accuracy, Jaccard Similarity (Jaccard), Mean Squared Error MSE, and Area Under the Curve (AUC).

\noindent\textbf{Datasets.} We use five datasets—MUTAG, NCI1, Tox21, Clintox, and FreeSolv—for evaluation. MUTAG is used for mutagenicity prediction, NCI1 for chemical compound classification, Tox21 for toxicity prediction, Clintox for drug-related toxicity, and FreeSolv for solvation energy prediction. We randomly select 50 graphs in each dataset to evaluate the attack effect and the rest for MGAE training.

\begin{figure*}[ht]
\setlength{\abovecaptionskip}{-5pt}
\begin{center}
\centerline{\includegraphics[width=\linewidth]{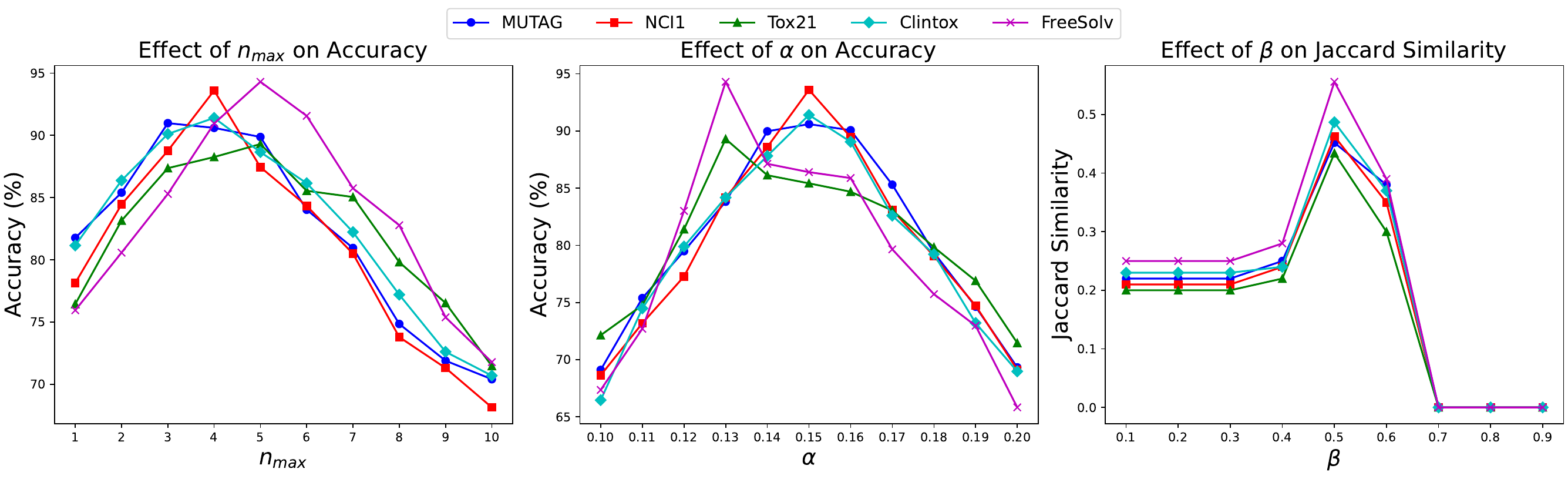}}
\caption{Effect of hyperparameters $n_{max}$,$\alpha$,and $\beta$ on Accuracy.
}
\label{combine}
\end{center}
\vspace{-0.8cm}
\end{figure*}

\noindent\textbf{Baselines.} We conduct baseline comparisons using four methods: direct application of DLG\cite{Zhu19}, DLG with binary projection (DLG+BP), iDLG\cite{Zhao20} (which predicts labels by applying a gradient-based linear classification layer on top of DLG, with positive and negative gradients for label prediction, a method also used in our approach), and iDLG with binary projection (iDLG+BP). We use a two-layer GCN model with a hidden channel dimension of 100, and the learning rate is set to 0.01. All methods are optimized using the Adam optimizer for 1000 attack iterations.

\begin{figure}[t]
\begin{center}
\centerline{\includegraphics[width=\linewidth]{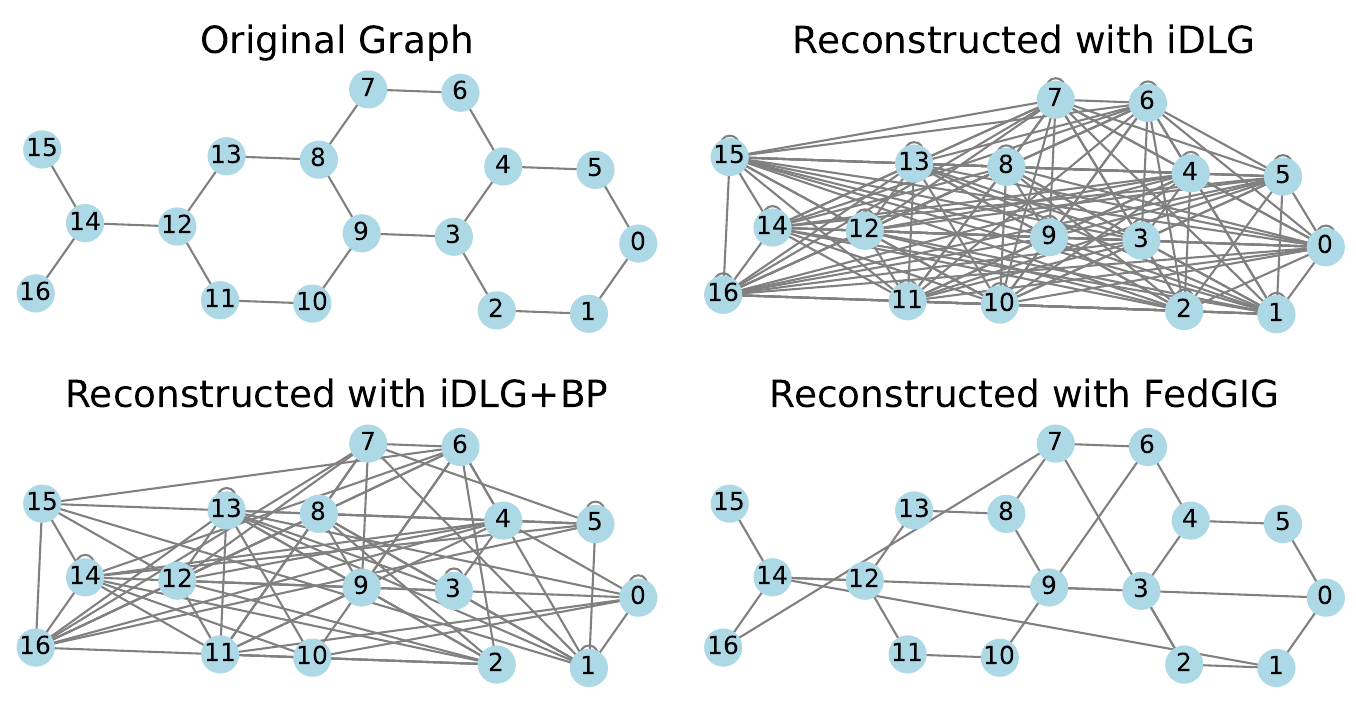}}
\caption{Comparison of graphs reconstructed with different methods.
}
\label{reconstruction}
\end{center}
\vspace{-0.8cm}
\end{figure}

\begin{table}[ht]
\caption{Ablation study of FedGIG with and without different components.}
\centering
\renewcommand{\arraystretch}{1.0} 
\begin{tabular}{l@{\hskip 5pt}c@{\hskip 5pt}p{1.4cm}@{\hskip 5pt}p{1.4cm}@{\hskip 5pt}p{1.4cm}@{\hskip 5pt}p{1.4cm}} 
\toprule
\multirow{2}{*}{\textbf{Method}} & \multicolumn{4}{c}{\textbf{Metrics}} \\
\cmidrule(r){2-5} 
& Accuracy $\uparrow$ & Jaccard $\uparrow$ & MSE $\downarrow$ & AUC $\uparrow$ \\
\midrule
FedGIG w/o AMC & 0.000 \tiny{$\pm$0.000} & 0.098 \tiny{$\pm$0.002} & 0.688 \tiny{$\pm$0.014} & 0.511 \tiny{$\pm$0.011} \\
FedGIG w/o SR &0.821 \tiny{$\pm$0.002} & 0.314 \tiny{$\pm$0.008} & 0.273 \tiny{$\pm$0.020} & 0.601 \tiny{$\pm$0.015} \\
FedGIG & 0.936 \tiny{$\pm$0.002} & 0.463 \tiny{$\pm$0.012} & 0.103 \tiny{$\pm$0.002} & 0.679
\tiny{$\pm$0.013} \\
\bottomrule
\end{tabular}
\label{table:ablation}
\end{table}



\subsection{Performance Comparisions}
As summarized in Table \ref{table1} and Figure \ref{reconstruction}, our method FedGIG outperforms other baseline GIA methods across five datasets and four metrics. The DLG and iDLG methods, designed for continuous image and natural language vector data, fail to correctly restore any discrete values in the adjacency matrix during the attack. The label inference method added to iDLG over DLG offers little help in improving the similarity of the adjacency matrix in single-task classification scenarios with a small number of categories. Still, it provides significant assistance in multi-task classification scenarios like Tox21. The DLG+BP and iDLG+BP methods, which incorporate a discretization projection of continuous values post-optimization, show a substantial increase in the similarity of the restored adjacency matrix to the true graph structure. However, they cannot eliminate the ghost edges present in the optimization and fail to ensure the undirected nature, symmetry, and sparsity of the graph, resulting in a significant gap from the true graph structure. The optimized graph structures remain overly dense, with many edges that do not exist in reality. Our FedGIG method applies continuous and gradually tightening constraints targeting the properties of graph structures during the optimization process and is adept at learning basic and common subgraph structures on molecular graphs with similar properties. This approach further significantly enhances the accuracy of the restored graph structures.

\subsection{Effects of Hyperparamters}
We performed a grid search over all hyperparameters to find the optimal value for each parameter. As can be seen from the left subgraph of Figure \ref{combine}, for the constraint on the maximum degree of each node, $n_{\text{max}}$, the best reconstruction accuracy was achieved at $n_{\text{max}} = 4$ for MUTAG, NCI1, and Clintox. The Tox21 and FreeSolv datasets, which have nodes with higher degrees, achieved optimal values at $n_{\text{max}} = 5$. The middle subgraph of Figure \ref{combine} indicates that the contraction rate of the constraint on the number of edges per node, $\alpha$, is optimal around 0.15 for MUTAG, NCI1, and Clintox. The Tox21 and FreeSolv datasets, due to their nodes having a larger number of connections, should not contract too quickly; otherwise, they may prematurely converge to local optima during optimization. Therefore, the optimal value is around 0.13. The right subgraph of Figure \ref{combine} shows that the threshold for the discretization projection of the adjacency matrix, $\beta$, is best at 0.5.

\subsection{Ablation Studies}
As shown in Table \ref{table:ablation}, we conduct ablation studies by comparing FedGIG without adjacency matrix constraining (FedGIG w/o AMC), FedGIG without subgraph reconstruction (FedGIG w/o SR) and FedGIG. The result shows that adjacency matrix constraining serves as the fundamental module in FedGIG. When employed independently, it can recover a substantial portion of the graph. However, in the absence of adjacency matrix constraining, which enforces symmetry, sparsity constraints, and binary discretization projections, the subgraph reconstruction module fails to achieve optimal reconstruction performance. This limitation can be attributed to numerous dense continuous values within the graph, which hinders the MGAE from effectively capturing the local subgraph patterns necessary for accurate reconstruction. Consequently, the subgraph reconstruction module must be applied after the adjacency matrix constraining module to ensure enhanced reconstruction accuracy.

\section{Conclusion} 
This paper introduces FedGIG, a novel method for more precise gradient inversion attacks in FGL. FedGIG's effectiveness stems from its two core modules: adjacency matrix constraining and subgraph reconstruction. The former optimizes the graph structure reconstruction by enforcing symmetry, sparsity, and discreteness, while the latter employs a masked graph autoencoder to recover local graph patterns accurately. Extensive experiments demonstrate that these modules significantly enhance the accuracy of graph data recovered by FedGIG over existing methods.

\bibliographystyle{IEEEbib}
\bibliography{icme2025references}

\vspace{12pt}
\color{red}

\end{document}